\documentclass[12pt,journal,compsoc]{IEEEtran}

\ifCLASSOPTIONcompsoc

  \usepackage[nocompress]{cite}
  \usepackage{graphicx}
\else

  \usepackage{cite}
\fi

\usepackage{setspace}
\usepackage{caption}
\usepackage{float}
\usepackage{booktabs}
\usepackage[font=small,skip=0.8pt]{caption}
\usepackage{hyperref}
\usepackage{arabtex}
\usepackage{utf8}
\setcode{utf8}
\usepackage{placeins}
\usepackage{ragged2e}
\usepackage{footnote}
\makesavenoteenv{table}
\makesavenoteenv{tabular}
\begin{document}
\title{The Challenges of Persian User-generated Textual Content: A Machine Learning-Based Approach}
\author{Mohammad Kasra Habib\\   Stuttgart University \\ISTE/Empirical Software Engineering\\
E-mail: kasra.habib@iste.uni-stuttgart.de }

\markboth{Stuttgart University}%
{Shell \MakeLowercase{\textit{et al.}}: Bare Advanced Demo of IEEEtran.cls for Journals}
\IEEEdisplaynontitleabstractindextext

\IEEEtitleabstractindextext{
\begin{abstract}
\justifying
Over recent years a lot of research papers and studies have been published on the development of effective approaches that benefit from a large amount of user-generated content and build intelligent predictive models on top of them. This research applies machine learning-based approaches to tackle the hurdles that come with Persian user-generated textual content. Unfortunately, there is still inadequate research in exploiting machine learning approaches to classify/cluster Persian text. Further, analyzing Persian text suffers from a lack of resources; specifically from datasets and text manipulation tools. Since the syntax and semantics of the Persian language is different from English and other languages, the available resources from these languages are not instantly usable for Persian. In addition, recognition of nouns and pronouns, parts of speech tagging, finding words' boundary, stemming or character manipulations for Persian language are still unsolved issues that require further studying. Therefore, efforts have been made in this research to address some of the challenges. This presented approach uses a machine-translated datasets to conduct sentiment analysis for the Persian language. Finally, the dataset has been rehearsed with different classifiers and feature engineering approaches. The results of the experiments have shown promising state-of-the-art performance in contrast to the previous efforts; the best classifier was Support Vector Machines which achieved a precision of 91.22\%, recall of 91.71\%, and \textbf{$F_1$} score of 91.46\%.
\end{abstract}

}

\maketitle

\IEEEdisplaynontitleabstractindextext

%
\IEEEpeerreviewmaketitle

\IEEEraisesectionheading{\section{Introduction}\label{sec:introduction}}

\IEEEPARstart{R}{ecently}, structured and unstructured user-generated content throughout the internet has been dramatically increased. Unstructured data can be easily perceived and analyzed by humans but are very hard for machines to understand. Likewise, extracting what other people think out of the generated content is an important task for decision-making \cite{Pang:2008:OMS:1454711.1454712} in business, politics, or beyond these domains.\\

Utilizing and analyzing raw user-generated content and deriving models based on them makes it more valuable, as per the The-Guardian \cite{lewis_pegg_hern_2018} : \lq\lq Derivatives of data (from user-generated content), which includes predictive models, or clusters of the population in psychological groupings, can be highly valuable to companies involved in micro-targeting advertisements to voters\rq\rq ; involving politics is beyond the scope of this research. \\

All types of data (i.e., including images, text, or videos), which is created by users of an unknown system or service on the internet, is called to be user-generated content \cite{moens2014mining, chen2018modeling}. After all, users have different languages, and the textual contents can be generated with different syntax and semantics. A huge amount of researches has been conducted for analyzing languages such as English \cite{vaziripour2016analyzing} and is still actively continuing.\\

Moreover, this research targets analysis of Persian user-generated textual content and the challenges that come with it, regrettably, there is an inadequate number of researches in exploiting machine learning approaches for the Persian language. On the other hand, analyzing Persian textual content also suffers from a lack of resources \cite{sarrafzadeh2011cross, shamsfard2011challenges} such as datasets and text manipulation tools.\\

Since the syntax and semantics of Persian are different from languages such as English, recognition of nouns and pronouns, parts of speech tagging, finding words' boundaries, stemming, or character manipulations are also different and are still unsolved issues that require further studying.\\

Therefore, efforts have been made in this research to address the main challenges. This presented approach conducts a case study of sentiment analysis for the Persian language. The results of this empirical approach have shown promising state-of-the-art performance in contrast to the previous efforts. Assuredly, this research makes the following contributions:

\begin{enumerate}
\item A dataset
\item In-depth identification of the main challenges in applying machine learning to Persian text
\item Presenting a state-of-the-art performance for Persian text classification 
\item Demonstrates tools impotence for preprocessing Persian text
\item Exhibits that classical machine learning approaches can proffer similar or sometimes even better performance to neural networks at the presence of an ideal dataset
\end{enumerate}

\section{Related Works}
Although the applicability of machine learning in natural language processing is extensively studied by scholars for some languages such as English, nevertheless, some suffer lacking it.\\

As of related work, this work picks the most relevant ones to its case study for comparison, since some effort have been devoted to other applications of machine learning to Persian.\\

Elham et al. \cite{vaziripour2016analyzing} question whether they can automatically analyze the sentiment of individual tweets in Persian. Their goal is to determine the individual tweets changing sentiment over time concerning the number of trending political topics. They conclude the challenges of their work in three cases:
\begin{enumerate}
\item lack of a sentiment lexicon and part-of-speech taggers, 
\item frequent use of colloquial words,
\item and, unique orthography and morphology characteristics.
\end{enumerate} 

For this work, they have collected over 1 million tweets of political domains in the Persian language, with an annotated dataset of over 3,000 tweets.
They deployed Naive Bayes and Support Vector Machines. Based on their finding, SVM outperformed Naive Bayes with an average accuracy of 56\% and as higher as 70\%.\\

Ehsan Basiri et al. \cite{basiri2017sentence} addresses the problems that come with sentiment analysis and builts three new resources, SPerSent, which contains customers' comments from the Web, CNRC, a lexicon corpus, and a new stop-word list.\\

Finally, they evaluate the resources with Naive Bayes. They conclude, \lq\lq the performance, with regard to all evaluation measures, are better when CNRC is used as the lexicon for labeling the SPerSent\rq\rq. The best-observed precision, recall, and \textbf{$F_1$} score are 92\%, 87\%, and 89\% respectively.\\

An important step for any machine learning related task is feature engineering. The following papers took a step forward to investigate the impact of different feature engineering methodologies for Persian text.\\

Ayoub Bagheri et al. \cite{bagheri2014persian}, investigated four feature selection approaches for sentiment classification; Document Frequency; Term Frequency Variance; Mutual Information; and Modified Mutual Information.
Next, Naive Bayes is fit to evaluate features' performance. The highest score is attained with Modified Mutual Information features. The final precision, recall, and \textbf{$F_1$} score are 90.72\%, 85.26\%, and 87.84\% each respectively.\\

Kia Dashtipour et al. \cite{dashtipour2017comparative}, proposes a novel sentiment analysis framework for the Persian language. Different feature engineering and their combinations are evaluated. As a result, the combination of unigram, bigram, and trigram presented the best performance with succeeding 88.36\% accuracy.\\

Besides traditional machine learning algorithms for natural language processing, one can apply neural networks to Persian text.\\

Kia Dashtipour et al. \cite{dashtipour2018exploiting}, tend to exploit Deep Learning for Persian sentiment classification. They compared, state-of-the-art shallow MLP based machine learning model with deep autoencoders and deep CNNs. Finally, the proposed CNNs model presents better performance than MLP and autoencoders with an achieved precision, recall, and \textbf{$F_1$} score of 84\%, 83\%, and 83\% respectively.\\

Behnam Roshanfekr et al. \cite{roshanfekr2017sentiment}, studies neural networks to accomplish sentiment analysis for Persian text. They conclude that deep learning models outperform other models with a precision of 59.1\%, recall of 52.2\%, and \textbf{$F_1$} score of 55.4\%.

\section{The Challenges of Processing the Persian User-generated Textual Content}
Before discussing the challenges that come with Persian user-generated textual content, it is better to establish a basic understanding of the language.\\

Farsi-e-Dari\footnote{Dari means \lq\lq Darbari\rq\rq\ (which in English means the language of the royal court) \cite{christinenollekarimi}.} is known as Dari in Afghanistan; one of two official languages \cite{cia2018}, Farsi in Iran; the only official language \cite{anew}, Tajiki in Tajikistan; the only official language \cite{ciataj2018} and called Persian in the English language; which all refers to Farsi-e-Dari \cite{spooner2012persian}.  Each of these names (Dari, Farsi, or Tajiki) refer to a different accent of Persian, it is important to notice its vocabulary is shaped by its environment and broader culture. Yet still, a purer (less influenced) form is spoken by Afghans than other speakers \cite{spooner2012persian}. \\

This beautiful language belongs to the Indo-European language family \cite{spooner2012persian}. Historically, the extent this language spoken ranges from the borders of India in the east, Russia in the north, southern shores of the Persian Gulf to Egypt, and the Mediterranean in the west \cite{spooner2012persian}. Currently, Persian also understood in parts of Armenia, Azerbaijan, India, Iraq, Kazakhstan, Pakistan, Turkmenistan, Uzbekistan, China, and Turkey \cite{bbc2018}. Persian originates from the Great Khorasan \cite{alikuzai2011aryana} (which Afghanistan's major current Persian speaking territories formed the major portion of Khorasan \cite{alikuzai2011aryana}).

\subsection{Challenges in Adopting Tools Build for English and Arabic Language Processing}

Over recent years plenty of text processing tools are built for English \cite{vaziripour2016analyzing}. One might think exploiting them for Persian would be an advantage. Unfortunately, these tools are not adoptable due to the variance in their grammar, syntax, and semantics; 
\begin{itemize}
\item  one can notice the syntax as the biggest difference, e.g., Persian is right-to-left, where English is left-to-right;  
\item  next, parts of speech tagging is another; 
\item  and ambiguity in word morphology and character manipulation are another barrier to be considered.\\
\end{itemize}
On the other hand, applying tools build for Arabic appears a good option, since this language adopts the Arabic character set and add four more to it \cite{shirali2010arabic}. However, both languages might look similar when it comes to writing, yet they are two different languages, and the syntax and semantics of both languages are different. \\

Hence, the Persian language vocabulary is exposed by Arabic grammar. For example, the words with the Arabic root bring irregular plural forms, while Persian uses a suffix to build plural forms \cite{anew}. Thereupon, not to forget Arabic language suffers from lack of tools and research like Persian. Besides, the influence do not imply that the tools built for Arabic text processing are instantly usable for Persian. Even they add-up to the complexity of the language, which will be discussed in the forthcoming sections.\\

All in all, the differences between Dari, Arabic, and English languages cause to engineer or develop new tools from scratch.

\subsection{{Challenges in Persian Text Processing}\label{sec:darichallenges}}

The written structure of Persian itself is complex than the languages like English. For instance, the appearance of the homographic word (the once which look alike, but have different meanings\footnote{The word \lq\lq\<شانه>\rq\rq\ ---/Sh\^ana/ can mean \lq\lq Shoulder\rq\rq\ or \lq\lq Comb\rq\rq\ depending its usage at the sentance, there are many examples of this condition.}) and use of irregular (which comes from Arabic) and with-suffix plural form which needs to be addressed \cite{shamsfard2011challenges, baluch2006persian}. \\

Moreover, there are many suffixes, prefixes, pronouns, and other parts that can be written separately or connected, which all are open to further research \cite{anew}. However, there are some research conducted to apply machine learning to Persian text, which are not adequate.\\

The main challenges to exploit machine learning models to process Persian text are concluded in lack of resources, ambiguities in character manipulation, morphology, identifying words boundary, and syntax analysis.

\subsubsection{The Challenges of Character Encoding}
Frequently one can think that Persian is a variant of the Arabic language. It is explicit that Persian and Arabic are two distinct languages, even they belong to different language families. Therefore, it is natural that they have some similarities in terms of syntax by the cause of alphabet adoption.\\

Fundamentally, computers deal with numbers. They store letters and other characters by assigning a number to each of them \cite{whatisunicode}. Before the invention of Unicode, there were hundreds of systems \cite{whatisunicode} and finding a unique way to rep- resent information was relatively difficult \cite{qasemizadeh2014challenges}. Resuscitation of Unicode is an effort to software internationalization, especially on the Web. Unicode system is designed to assign one unique code for each character even if the character is used in multiple languages \cite{esfahbod2004persian}. Persian scripts are written based on Arabic characters ($\langle$U+0600—U+06FF$\rangle$ block) with some extra and modified characters \cite{whatisunicode, esfahbod2004persian}. The current Unicode framework for Persian is insufficient \cite{qasemizadehtranscription}.\\

It is important to know that the design principle in Unicode to represent the relevant shapes are characters not glyphs \cite{esfahbod2004persian}. In Persian or Arabic a character can take four different shapes (glyph) depending on their position in the sequence (Table 1).\\

\begin{table}[h]
\centering
\caption{Persian character shapes}
\includegraphics[width=8cm]{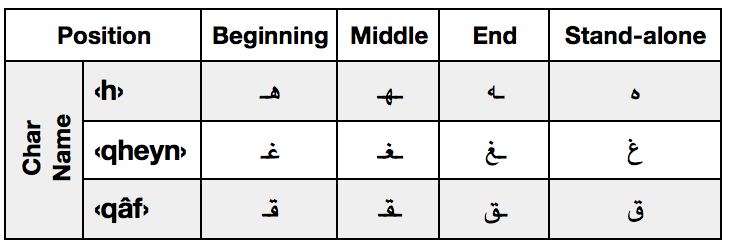}
\end{table}

It is noteworthy that for each four visual form (glyph) of a character there is only one single code. Therefore, an algorithm is charged to handle four visual form of a character in a sequence \cite{odeh2012steganography, alshahrani2017hybrid}. This algorithm attaches special characters such as Zero Width Joiner (ZWJ), Zero Width Non-Joiner (ZWNJ) and Right- to-Left Override (RLO) \cite{qasemizadehtranscription}.\\

Take ZWNJ for instance. Using it after a code means that the character before ZWNJ must appeare in one of its final forms (glyph), a character after ZWNJ forces the character to appeare in one its initial forms (glyph), and similarly characters after RLO should be represented as strong right-to-leftt character \cite{qasemizadeh2014challenges}. There are also other standards proposed to use for character representation, e.g., ISIRI 6219:2002 (usual in Iran). Despite these standards, Persian keyboard layout is using different codes and many of Persian users do not use the same encoding standards \cite{qasemizadeh2014challenges}. In addition, using different encodings paves the way for more challenges.\\

Furthermore, if one is asked to write the plural form of \lq\lq\< بخش>\rq\rq\ (which means \lq\lq Section\rq\rq\ in English),  a suffix \lq\lq\<ها>\rq\rq\ ---/h\^a/ will be added at the end of the word. Therefore, the plural forms based on different standards representing the words deferentially as follow:\\

\begin{arabtext}
\hspace{1.1cm} \<بخش>+\LR{SPACE}+\<ها>=\<بخش ها> \hspace{0.1cm}\LR{(1)}\\
\hspace{1.7cm} \<بخش>+\LR{ZWNJ}+\<ها>=\<بخش ها>\hspace{0.27cm}\LR{(2)}\\
\hspace{1.7cm} \<بخش>+\<ها>=\<بخشها>\hspace{2.5cm}\LR{(3)}\\ \vspace{1cm}
\end{arabtext}

Moreover, such examples in the corpus effects on measuring precision and recall for classification or clustering while the same feature can be represented as two or more different vectors when word frequencies are calculate \cite{qasemizadeh2014challenges}.

\subsubsection{Ambiguities in Character Manipulation}
Starting Persian text analytics means tackling with lots of challenges. It is quite possible to input Arabic characters instead of the standard Persian ones \cite{qasemizadeh2007farsi}. A common mistake which usually happens is utilizing of \lq\lq\<ی>\rq\rq\ instead of \lq\lq\<ي>\rq\rq\ or \lq\lq\<ك>\rq\rq\  instead of \lq\lq\<ک>\rq\rq.\ This mix-up causes a problem while one looks dictionaries or calculates word frequencies whereof different encoding strings.  On the other hand, even if one is used to input this (combination of mixed Unicode characters) on the Google search engine, that one can end up with different results, since pages are ranked based on different words (the same appearing word with different Unicode in background are treated as different words) \cite{qasemizadeh2014challenges}.\\

Another character which causes the same problem is short vowels. A short vowel in Persian transcriptions never appears alone \cite{anew}. If one is used, then they will be coded independently \cite{qasemizadeh2014challenges}, which can also raise the problem of same appearing word width different Unicode.\\

Moreover, other problem of this kind can happen in regards to using TATWEEL character; a visual character which helps Persian and Arabic words to appear in different widths \cite{ibrahim2002characteristics}. To tackle this challenge, \cite{qasemizadeh2014challenges} proposes to build a standardized procedure such us using a mapping between Persian and Arabic characters.

\subsubsection{Ambiguity at Words' Boundary}
Tokenization as part of preprocessing for text classification or clustering directly effects on the performance of machine learning algorithm. To convert documents into tokens one should simply find the word boundaries.\\

Tokenization of Persian documents are challenging due to different usage of delimitations; for example, Persian compound and light words are written in delimited form with ZWNJ. Besides, one can use the space character to form these words, which is not respected by users; even by the official organizations \cite{qasemizadeh2014challenges}. Furthermore, to tokenize defining space or ZWNJ as a boundary are not adequate. \\

Basically, one can think of using four visual forms (beginning, middle, end, stand-alone) as word boundaries, which the final form is a strong indication to end of a word. However, \cite{megerdoomian2000processing, hassel2004farsisum} shows that this technique with the Unicode system is not applicable.

\subsubsection{Ambiguity in Morphology}
Morphological ambiguities can arise based on two reasons \cite{qasemizadeh2014challenges}, (1) homograph words, and (2) word boundaries. \\

Take (1) for instance. The word \lq\lq\<مهر>\rq\rq\ can have different pronunciation and meanings with respect to the usage of short vowels which does not appear in the written text: with one usage of short vowels the word \lq\lq\<مهر>\rq\rq\ means Love, with another means to Seal and is used to indicate Mahr\footnote{In Islam, Mahr is an arbitrary payment, in the form of money or possessions paid by groom, to the bride at the time of marriage to appreciate her \cite{oman2010bargaining, blenkhorn2002islamic, freeland2000islamic}.}.\\

For (2) remember the example from section 3.2.1; that how a word can be treated as 3 (more than 3 are possible) different words due to Unicode mix-ups.\\

Therefore, similar problems can arise when lexical elements such as preposition, postposition or conjunctions appear separately or attached \cite{perry1985language, qasemizadeh2014challenges}. A solution to this challenge would be to follow the official Persian's orthography which recommends writing them separately, which is hard to guarantee. Therefore, a promising solution is to build a text normalizer.

\subsubsection{Ambiguity While Detecting Proper Nouns}
Importance of using part-of-speech tagging in text analytics is obvious. To tag nouns and pronouns, Arabic transcripts do not enjoy capitalization like English. Therefore, this characteristic is inherited by Persian's Arabic character adoption.\\ 

To solve this challenge, \cite{steinbach2000comparison} offers some heuristics to distinguishes proper-nouns from nouns.

\subsubsection{Ambiguity Syntax Analysis}
Another important ambiguity arises when one wants to construct possessives \cite{qasemizadeh2014challenges}. To construct possessives a short vowel $/e/$ is used, which does not appear in the writing text. \cite{qasemizadeh2014challenges} recommends to add this short vowel in writing scripts; for this regards as it is discussed in subsection 3.2.2, adding this short vowel can cause challenges for character manipulation.\\

Mainly the problems which may occur while Persian text analytics can be summarized to the inconsistency in its character representation and special orthography.\\

Furthermore, if one does not consider the mentioned challenges while applying machine learning to Persian transcripts it is quite possible to achieve unsatisfied results, which are far away from the expected performance. To remove the ambiguities, \cite{shamsfard2011challenges} proposes to use a combination of orthography and use standards which are defined in \cite{qasemizadeh2014challenges}. Handling the challenges laid in Persian scripts are the major obstacles of this research.

\subsubsection{FEnglish}
Another challenge to apply machine learning to Persian transcripts is that one can use English alphabets to write Persian words pronunciation. This style of writing is called \textit{FEnglish} which became very usual in social networking platforms. There are some tools\footnote{The following links are the tools for mapping FEnglish to Persian text: \url{http://www.dictionary-farsi.com/pinglish.asp}, \url{http://syavash.com/portal/pinglish2farsi/convertor-en}, \url{https://lingojam.com/FarsitoFinglish}} to convert the written pronunciations from English to Persian alphabet. This conversion is insufficient for textual analysis since it is just a mapping between English to Persian characters. For example, there are two or more characters which can generate almost the same phoneme (e.g., \lq\lq e\rq\rq\ and \lq\lq a\rq\rq) and can be used to write a Persian word with it. Since it is not an officially writing standard; writing a word with different selected characters can differ from one to another. \\ 

Therefore, this can cause to form a high dimension feature space. Yet still, this is considered an open challenge and needs further attention.\\

From the aforementioned sections, it is now clear that Persian has a complex orthography structure. This complex structure attaches more challenges to the generally available and the laid challenges in social networking platforms. Therefore, to achieve better performance one must consider these challenges.

\section{Experiments and Results}
This research uses a machine translated dataset, i.e., observations are originally written in English language which each is an assortment of reviews form Rotten Tomatoes. This dataset is originally collected by Pang and Lee \cite{pangb2005exploitingclassrelationshipsforsentimentcate}, for their work on sentiment treebanks, and likewise utilized by Socher et al. \cite{socher2013recursive} for Amazon's Mechanical Turk to create fine-grained labels for all parsed phrases into corpus. This research considers a bulk of only positive and negative observations. Therefore,  the dataset size is shrunken to 16278 instances which 3256 are kept for test purpose.

\subsection{Preprocessing and Feature Engineering}
Each instance form the culled dataset is applied a four step cleaning; unnecessary characters removal; normalization and stemming; TF-IDF is applied to assign equal weight to more frequent tokens; PCA is used for dimension reduction to keep 0.99\% of explained variance ratio to boost the training process.\\

Once the preprocessing step is satisfied, vector features are built based on word n-gram (i.e., where n = 1, 2 and 3) and character n-gram (i.e., n = 1) feature extraction methods. Furthermore, word n-gram models are famous for preserving order which a word appears in a document and if one is required to capture a deeper meaning, i.e., morphological makeups \cite{kulmizev2017power,  lesher1999effects, dipl134} shall go with character n-gram model.\\

Subsequently, Logistic Regression (LG) is fit to apprehend a baseline. Clearly, it can be inferred that the trained model with unigram features (word n-gram = 1) has imperceptibly higher performance than the rest (TABLE 1).


\begin{table}[h]
\centering
\caption{Logistic Regression's performance measurement on different feature extraction methods}
\resizebox{\columnwidth}{!}{
\begin{tabular}{lccc@{}}
\toprule
 & \multicolumn{1}{l}{\textbf{Precision (\%)}} & \multicolumn{1}{l}{\textbf{Recall (\%)}} & \multicolumn{1}{l}{\textbf{$F_1$ Score (\%)}} \\ \midrule
\textit{\textbf{Word 1-gram}} & 88.12 & 90.69 & 89.39 \\
\textit{\textbf{Word 2-gram}} & 80.46 & 94.63 & 86.98 \\
\textit{\textbf{Word 3-gram}} & 70.51 & 98.45 & 82.17 \\
\textit{\textbf{Char 1-gram}} & 64.87 & 76.41 & 70.17 \\ \bottomrule
\end{tabular}
}
\end{table}

Since the model's score comes from training set they hold suspect to be likely to overfitting. Additionally, unigram and bigram features have approximately similar performance. To make sure which feature set to select learning curves for all features are plotted (Fig. 1).

\begin{figure}[h]
\centering
\includegraphics[width=9cm]{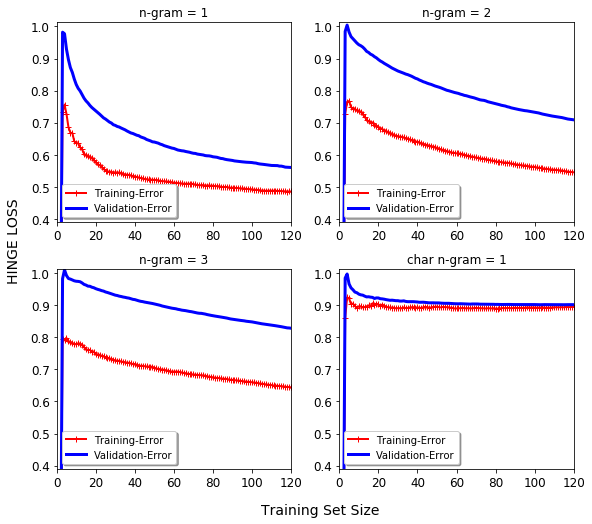}
\caption{Learning Curves}
\end{figure}

From Fig. 1, it easy to construe that word bigram and trigram features have a high variance (overfitting); there is a gap between each' two curves. It means the models are significantly admirable on the training set than the validation set. The model with character unigram is prone to underfitting. Therefore, the winner is the word unigram model.\\

Eventually, Random Forest Classifier is trained to select features; each feature is selected based on its mean weight, where each node's weight is equal to the number of training samples associated with it \cite{sklearn_api}. It transpires that the feature selection step did not improve the performance, which is dropped from the preprocessing pipeline.\\

Finally, KMeans is fitted as a preprocessing step, and to extract new features. To achieve the best number of related clusters, the Basian Gaussian Mixture model was used, which resulted in 37 clusters.  Three new feature sets are built;
\begin{enumerate}
\item \textbf{\textit{Distances}}: replaced instance with their distances to these 37 clusters; 
\item \textbf{\textit{Centers}}: instances replaced with their cluster centers; 
\item \textbf{\textit{Combined}}: the combination of two previous feature assortments and word unigram features (TABLE 2).
\end{enumerate}
\begin{table}[h]
\caption{Logistic Regression's performance measurement on extracted features}
\resizebox{\columnwidth}{!}{
\begin{tabular}{@{}lccc@{}}
\toprule
 & \multicolumn{1}{l}{\textbf{Precision (\%)}} & \multicolumn{1}{l}{\textbf{Recall (\%)}} & \multicolumn{1}{l}{\textbf{$F_1$ Score (\%)}} \\ \midrule
\textit{\textbf{Distances}}       & 68.54 & 86.98 & 76.67 \\
\textit{\textbf{Cluster Centers}} & 64.82 & 86.71 & 74.19 \\
\textit{\textbf{Combined}}        & 87.77 & 90.58 & 89.15 \\ \bottomrule
\end{tabular}}
\end{table}

From Table 2, it can be closed that the new features and even the combination of them with word unigram did not improved the performance; nothing astounding. Therefore, this research will remain with word unigram features. 

\subsection{Classifiers' Performance Study}
From the previous section, Logistic Regression with word unigram features showed a better performance. Hereabouts, this research will advance with training more models and fine-tuning the models' hyperparameters. \\

The chosen models are Logistic Regression (LG), SVM with a Stochastic Gradient Descent implementation (SGD SVM), Random Forest Classifier (RND), Linear Discriminant Analysis (LDA), and Multinomial Naive Bayes (MNB) which is suitable for classification with discrete features \cite{sklearn_api}.\\
%

Once classifiers' hyperparameter is fine-tuned with 10 fold of cross-validation (to save time, LDA was applied 3 fold cross-validation), it resembles that SVM (with SGD implementation and l2 regularization) shown a promising performance (Table 4). \\

Consequently, since the number of positive instances in the dataset is not scarce, plus an equivalent balance of precision and recall are required, this study presents ROC curve instead of precision and recall curves.\\

\begin{figure}[h]
\centering

\includegraphics[width=9cm]{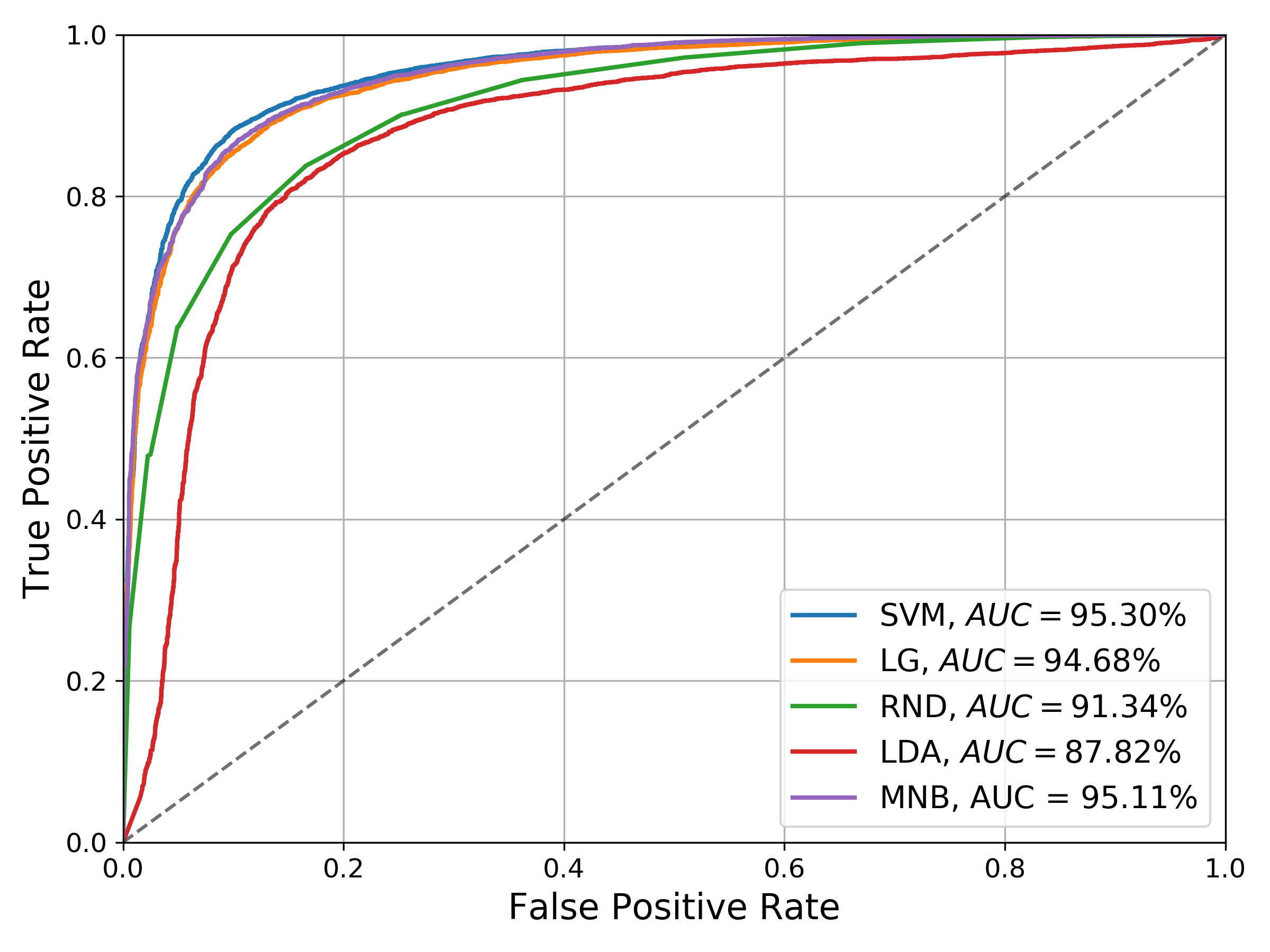}
\caption{Classifiers' ROC Curves}
\end{figure}

Eventually, this research ought to attempt ensemble learning; Soft Voting, Pasting, and AdaBoost classification models are studied. \\

First, all the previous algorithms are included for the Voting Classifier except a Gaussian Naive Bayes (GNB) is succeeded LDA considering it to wreck the performance. Second, an ensemble of Voting Classifiers are fitted; each is trained on 200 instances randomly sampled from the training set without replacement (Pasting). Finally, 100 Decision Tree Classifier (where max\_depth = 1) is trained to perform AdaBoost (Table 5).\\
\begin{table*}[!h]
\centering
\caption{Classifiers' performance measurement}
\resizebox{\textwidth}{!}{%
\begin{tabular}{@{}lcccccccc@{}}
\toprule
                     & \multicolumn{4}{c}{\textbf{Train Score}} & \multicolumn{4}{c}{\textbf{Test Score}} \\ \midrule
 &
  \textbf{Precision (\%)} &
  \textbf{Recall (\%)} &
  \textbf{$F_1$ Score (\%)} &
  \textbf{ROC AUC (\%)} &
  \textbf{Precision (\%)} &
  \textbf{Recall (\%)} &
  \textbf{$F_1$ Score (\%)} &
  \textbf{ROC AUC (\%)} \\
\textit{\textbf{SVM}} & 90.01    & 90.49    & 90.25    & 95.30   & 91.22    & 91.71    & 91.46   & 95.69   \\
\textit{\textbf{LG}}  & 88.12    & 90.69    & 89.39    & 94.68   & 89.06    & 92.20    & 90.60   & 95.11   \\
\textit{\textbf{RND}} & 86.79    & 83.76    & 85.25    & 91.34   & 89.09    & 84.45    & 86.71   & 92.64   \\
\textit{\textbf{LDA}} & 83.46    & 86.79    & 85.09    & 87.82   & 86.70    & 90.09    & 88.36   & 91.72   \\
\textit{\textbf{MNB}} & 84.76    & 93.85    & 89.07    & 95.11   & 86.51    & 94.47    & 90.32   & 95.62   \\ \bottomrule
\end{tabular}%
}
\end{table*}

Usually, training ensemble models provide better performance than training a single classifier. Nevertheless, even with ensemble learning, better performance is not always guaranteed.\\

Of Fig. 2, it is evident that SVM, LG, and MNB are the authoritative classifiers and are proffering a comparable performance. Thus, they are influencing the Voting Classifier's decisions. 
\begin{table}[h!]
\centering
\caption{Ensemble Learning's performance comparison}
\resizebox{\columnwidth}{!}{%
\begin{tabular}{@{}lcccccccc@{}}
\toprule
 &
  \multicolumn{2}{c}{\textbf{Precision (\%)}} &
  \multicolumn{2}{c}{\textbf{Recall (\%)}} &
  \multicolumn{2}{c}{\textbf{$F_1$ Score (\%)}} &
  \multicolumn{2}{c}{\textbf{ROC AUC (\%)}} \\ \midrule
                  & Train & Test  & Train & Test  & Train & Test  & Train & Test  \\ \cmidrule(l){2-9} 
\textbf{VTN}   & 87.58 & 89.29 & 93.59 & 94.42 & 90.48 & 91.79 & 95.87 & 96.47 \\
\textbf{PAS}  & 73.12 & 72.00 & 97.08 & 96.97 & 83.42 & 82.64 & 92.69 & 92.10 \\
\textbf{ADB} & 74.55 & 72.40 & 90.54 & 89.82 & 81.77 & 80.17 & 87.61 & 85.22 \\ \bottomrule
\end{tabular}
}
\end{table}

Moreover, among the applied ensemble classifiers, the Voting  classifier's scores are comparable with SVM (Fig. 3).\\

\begin{figure}[H]
\includegraphics[width=9cm]{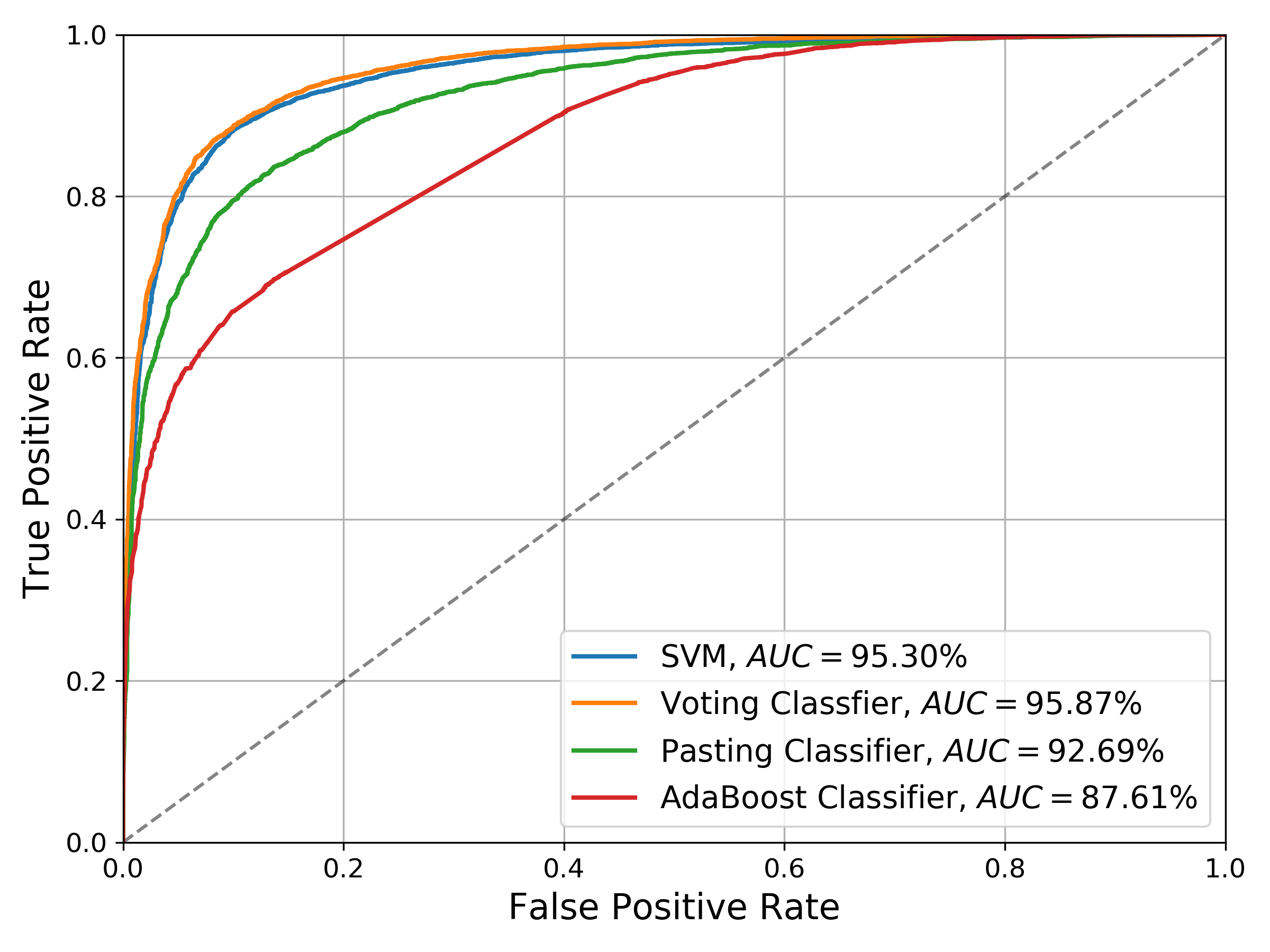}
\caption{ROC curves for SVM and Ensemble Learning}
\end{figure}

One can trade-off between precision and recall scores for the Voting Classifier to achieve nearly identical once to SVM (Fig. 3). Setting the decision threshold for recall to 0.91 will boost the precision to 90\% for the Voting classifier. \\

Yet, it is not enough, therefore supplementary (with respect to new threshold) ROC and AUC are calculated for this classifier, to make sure this model functions as good as SVM.


\begin{figure}[h!]
\includegraphics[width=9cm]{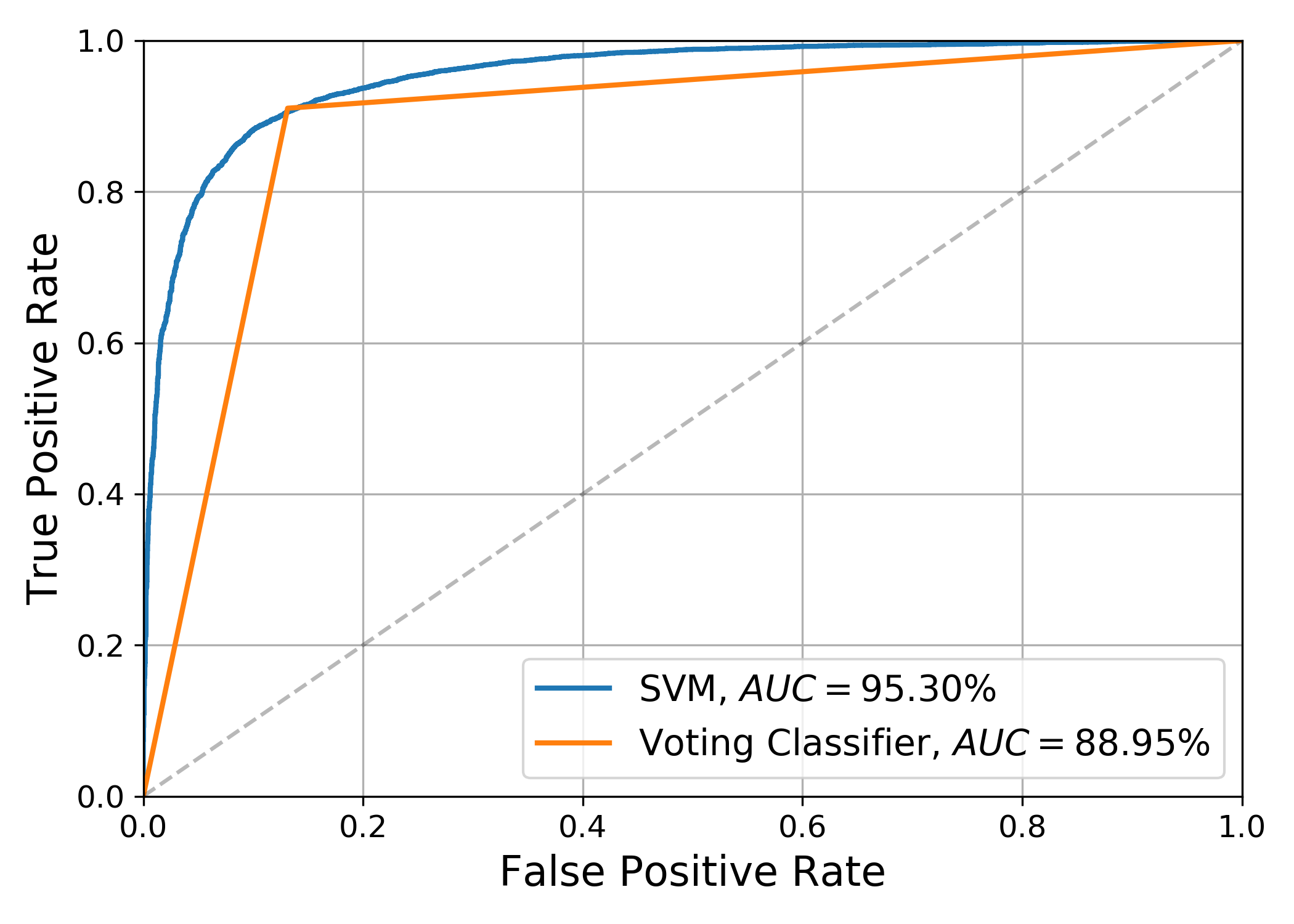}
\caption{SVM and Tweaked Voting Classifier's ROC curves}
\end{figure}
From Fig. 4, it is understandable that the curve is engineered and the tweaked model is not as shiny as SVM. Plus, contrasting its performance and its complexity, it does not deserve to replace SVM with it.\\

This research studied different feature engineering methods, classification algorithms, and ensemble learning. To sum up, SVM with word unigram features outperformed the rest, with an/a achieved/balanced precision, recall, and \textbf{$F_1$} score of 90\% on the train and 91\% on the test set.

\section{Evaluation and Future Work}
The achievment form the section 4, shown promising state-of-the-art performance in contrast to the previous efforts \cite{vaziripour2016analyzing, basiri2017sentence, bagheri2014persian, dashtipour2017comparative, dashtipour2018exploiting, roshanfekr2017sentiment} (Table 6).

\begin{table}[h]
\centering
\caption{Performance measurement comparison among this work and the related works}
\resizebox{\columnwidth}{!}{%
\begin{tabular}{@{}llcccc@{}}
\toprule
\textbf{References}                           & \textbf{Model}                       & \textbf{Precision (\%)} & \textbf{Recall (\%)} & \textbf{$F_1$ Score (\%)}  \\ \midrule
\textbf{This Study}            & SGD SVM                          & \large 91.22          & \large91.71       & \large91.46          \\
\textbf{E. Basiri \small et al.}\cite{basiri2017sentence}       & NB\footnote{Naive Bayes}                & \large 92.00             & \large 87.00          & \large 89.00                    \\
\textbf{A. Bagheri \small et al.}\cite{bagheri2014persian}      & NB                  & \large 90.72          & \large 85.26       & \large 87.84                   \\
\textbf{Kia D. \small et al. }\cite{dashtipour2018exploiting}    & CNN\footnote{Convolutional Neural Networks} & \large 84.00             & \large 83.00          & \large 83.00                   \\
\textbf{Behnam R. \small et al. }\cite{roshanfekr2017sentiment} & NN\footnote{Neural Networks}               & \large 59.10           & \large 52.20        &\large 55.40                  \\
\bottomrule
\end{tabular}
}
\end{table}
E. Basiri et al. \cite{basiri2017sentence}, confers a  precision score of 0.92; a slight difference of 0.0078 in contrast to this work. On the other hand, it presents a lower recall score, which is 0.87 (i.e., it can not detect  13\% of positive instance) in contrast to this work, which is 0.91. A convenient method to compare two classifiers is to embed their precision and recall scores in one metric called \textbf{$F_1$} score (the \textit{harmonic mean}). Therefore, looking at the \textbf{$F_1$} score of both studies, it is clear that this research outperforms E. Basiri et al. \cite{basiri2017sentence} with a difference of 2.46\% while trading-off between the precision and recall.\\

Differently, Elham et al. \cite{vaziripour2016analyzing}  and Kia Dashtipour et al. \cite{dashtipour2017comparative} provides only the Accuracy score instead of precission, recall and \textbf{$F_1$} score, which is not the preferred metric of evaluation. Thus, this work exhibits a higher (i.e., 93\%) accuracy than the previous efforts.\\

Eventually, this research assumes that better performance can be reached if the following recommendations are satisfied:
\begin{enumerate}
\item Existence of an ideal dataset rather than (this) machine-translated one; mistakes in translation were observed in the dataset, as machine translation itself requires further study \cite{slocum1985survey}.
\item The existence of sophisticated tools for preprocessing is required; this research applied Hazm for stemming. Hence, Hamz is the state-of-the-art preprocessing tool for the Persian language it needs further improvements. Take for instance, the word \lq\lq\<ارام>\rq\rq\ (---which in English it means \lq\lq Quiet\rq\rq ) was wrong stemmed to \lq\lq\<ارا>\rq\rq\ (---which in English it means \lq\lq Vote\rq\rq) or the word \lq\lq\<بی نظير>\rq\rq\ was miss tokenized in two tokens \lq\lq\<نظير>\rq\rq\ and \lq\lq\<ب>\rq\rq .
\end{enumerate}
From the conducted research it is understandable that wielding the Persian text is a challenging responsibility. Furthermore, ensuing potential research for the future is to work on building rich resources and efficient preprocessing tools. The current solutions are based on traditional machine learning approaches, as well as conducting sentiment classification of Persian text with deep learning will be further fascinating to work in the future.

\section{Conclusion}
This work started by addressing the open challenges to apply machine learning to handle Persian user-generated textual content. Though there is plenty of support for English, unfortunately, adapting these resources is not a solution due to complexity in the syntactical and semantical structure of Persian language. Notwithstanding, several efforts are accomplished to develop preprocessing tools and employ machine learning to classify Persian sentiments, which are not adequate, therefore, in-depth studies are demanding.\\

First, this study applied a four-step preprocessing, features vectors are constructed based on word and character n-gram techniques. Later, Random Forest Trees are utilized for feature selection. Then KMeans is applied to create three new feature sets. From the results, it was apprehended that word unigram features without feature selection outperformed the rest.
Second, five classifiers (Support Vector Machines, Logistic Regression, Random Forest Classifier,  Linear Discriminant Analysis, and Multinomial Naive Bayes) and three ensemble learning methods (Voting, Pasting and AdaBoost) are trained and evaluated.\\

Finally, word unigram features and  SVM with gradient descent implementation outperformed the rest, with an achieved precision, recall, and \textbf{$F_1$} score of 90\% on the train and 91\% on the test set.


\bibliographystyle{IEEEtran}
\bibliography{./references}

\end{document}